\documentclass[conference]{IEEEtran}
\IEEEoverridecommandlockouts

\usepackage{cite}
\usepackage{amsmath,amssymb,amsfonts}
\usepackage{algorithmic}
\usepackage{graphicx}
\usepackage{textcomp}
\usepackage{xcolor}
\usepackage{booktabs}
\usepackage{verbatim}
\usepackage{subcaption}
\usepackage{xcolor}

\newcommand{\OrbitALIF}{\textsc{OrbitALIF}}

\newcommand{\OrbitANN}{\textsc{OrbitANN}}
\newcommand{\OrbitUnet}{\textsc{OrbitUNet}}

\graphicspath{{figures/}}

\begin{document}

\title{\OrbitALIF: An Efficient Spiking Federated Learning Framework for Onboard Cloud Removal}

\author{
  \IEEEauthorblockN{Bohan Zhang, Chenyu Xu, Yijie Mao, Yuanming Shi}
  \IEEEauthorblockA{
    School of Information Science and Technology, ShanghaiTech University, Shanghai, China\\
    Email: \{zhangbh2023, xuchy2024, maoyj, shiym\}@shanghaitech.edu.cn}\thanks {This work was supported by the National Natural Science Foundation of China under Grant 62571331, the work of Yuanming Shi was supported by the National Natural
    Science Foundation of China under Grant 62522117 and Grant 62271318.
    (\textit{Corresponding author: Yijie Mao})}
}
\IEEEaftertitletext{\vspace{-4mm}}

\maketitle

\begin{abstract}
  Low-earth-orbit (LEO) satellites enable high-resolution, large-scale Earth observation for applications such as disaster monitoring and environmental surveillance. However, cloud coverage often obscures the Earth's surface, and conventional cloud-removal pipelines that download cloudy images to ground stations for processing suffer from limited contact windows, constrained satellite-to-ground bandwidth, and high latency.
  In this work, we propose a novel satellite federated learning framework for cloud removal across LEO constellations, named orbital attention leaky integrate-and-fire (\OrbitALIF). \OrbitALIF\ performs both onboard training and inference using a compact $2.30$\,M-parameter spiking neural network (SNN) backbone with an adaptive gated fusion module (AGFM) and a spectral-spatial hybrid attention module (SHAM), combined with a decentralized federated learning strategy that shares model weights via inter-satellite links. Our experiments show that \OrbitALIF\ achieves competitive cloud removal quality while consuming only $0.287$\,mJ per inference on neuromorphic hardware, a $72.3\times$ ($98.6\%$) energy reduction versus an equivalent artificial neural network (ANN).

\end{abstract}


\section{Introduction}\label{sec:intro}

Low-earth-orbit (LEO) satellites have become an essential
infrastructure for Earth observation, supporting time-sensitive
applications such as disaster response~\cite{ref:earthobs2},
environmental monitoring~\cite{ref:earthobs1}, and large-scale
remote-sensing analytics. However, heavy cloud coverage can severely degrade the quality of the captured imagery.
Cloud removal, which aims to recover the original surface from a
cloud-contaminated observation, is therefore a key preprocessing step
for remote-sensing tasks.



Early cloud removal methods employed basic convolutional networks but only for thin cloud~\cite{Unet}. To alleviate the ambiguity caused by cloud occlusion, subsequent methods incorporated additional observations. SAR-assisted approaches such as DSen2-CR~\cite{dsen2} exploit co-registered cross-sensor information, while multi-temporal methods leverage repeated acquisitions of the same scene. Although effective, these approaches introduce additional requirements for auxiliary sensors, image registration, or repeated revisits. Other methods therefore remove the cloud directly from optical observation through powerful representation model such as MemoryNet~\cite{Memory}. Recently, diffusion-enhanced methods further improve cloud removal quality through iterative denoising~\cite{Cave,diffusion}, but at the cost of substantial computational complexity and inference latency. Consequently, existing cloud removal solutions face two fundamental deployment challenges. First, performing cloud removal on the ground requires cloud-contaminated images to be transmitted through bandwidth-constrained links, resulting in high latency. Second, sophisticated cloud removal models are often too computationally and energy intensive for practical onboard training and inference.


To address these challenges, satellite federated learning (SFL) has been proposed, leveraging the entire satellite constellation as a collaborative computing platform~\cite{SFL11}. By performing local training on each satellite and sharing model parameters via inter-satellite links (ISLs)~\cite{ref:fedmega2024,ref:dfedsat}, SFL enables access to richer and more diverse training data without transmitting raw imagery to the ground, while also preserving data security and privacy. However, prior SFL efforts have been limited to classification and detection tasks. The more challenging task of dense pixel-level cloud removal, which is a per-pixel regression problem under physically non-independent and identically distributed (non-IID) conditions, remains largely unexplored.

Moreover, SFL alone does not reduce the energy cost of onboard inference, which can exceed the power budgets of LEO satellites. This calls for an inherently energy-efficient model architecture.
Spiking neural networks (SNNs) offer a solution through event-driven inference~\cite{ref:roy2019}, consuming power only when spikes are emitted, and recent work has demonstrated their feasibility for low-level vision tasks such as deraining~\cite{ref:esdnet}. Yet conventional binary spikes lack sufficient expressiveness for dense per-pixel regression and aggressively reducing computation degrades cloud removal quality.


In this work, we propose the first onboard cloud removal framework named orbital attention leaky integrate-and-fire (\OrbitALIF) that jointly supports onboard training and onboard 
inference for cloud removal over a LEO satellite
constellation.By employing a decentralized SFL framework and an efficient SNN backbone, the proposed \OrbitALIF\ significantly reduces per-satellite energy consumption while maintaining competitive testing accuracy.

The main contributions of this paper are summarized as follows:

\begin{itemize}

  \item We propose \OrbitALIF\, an onboard SNN-based federated
        cloud-removal framework for LEO satellite constellations. To the best
        of our knowledge, this is the first work to go beyond classification
        tasks and combine onboard SNN inference with cross-orbit SFL for the low-level vision task of cloud removal.

  \item We design a lightweight SNN backbone for onboard cloud removal. Compared with existing cloud removal neural networks, the proposed backbone introduces two key architectural novelties. First, an adaptive gated fusion module (AGFM) replaces the rigid skip connections of a standard U-Net with a learnable gating mechanism. Second, a spectral-spatial hybrid attention module (SHAM) is introduced with two branches: a temporal attention branch that weights per-time-step importance, and a spectral attention branch that reweights frequency components via the 2-D fast Fourier transform (FFT) while preserving spatial phase structure.  The complete model
        contains only $2.30$\,M parameters, making it suitable for resource-constrained
        satellite deployment.

  \item We evaluate \OrbitALIF\ on the public CUHK-CR~\cite{diffusion} cloud
        removal benchmark under hierarchical inter-satellite links. \OrbitALIF\ achieves performance comparable to an ANN U-Net model of the same architecture with a $72.3\times$ reduction in inference energy, demonstrating the promise of spiking computation for onboard cloud removal.
\end{itemize}

\section{System Model}\label{sec:sysmodel}

\subsection{Satellite Communication Model}\label{sec:sysmodel-setup}

In this work, we consider a constellation consisting of $N$ orbital planes, each containing $K$ satellites. We denote the $j$-th satellite within the $i$-th orbital plane by $s_{i,j}$. Satellites are assumed to communicate with one another via ISLs. Two decision metrics are introduced to determine whether a pair of satellites can establish communication:

\textit{1) Line-of-sight (LoS) visibility}:
Let $s_{i,j}$ and $s_{i',j'}$ be an arbitrary satellite pair (hereafter denoted simply as $s_1$ and $s_2$ when no ambiguity arises), with altitudes $h_1$ and $h_2$ above the Earth's surface, respectively. The horizon half-angle for satellite $s_k$ is $\rho_k = \arccos\bigl(r_E/(r_E + h_k)\bigr)$, which determines the angular radius of the cone within which the satellite can establish LoS with other spacecraft. Let $\theta_k^{v}$ and $\theta_k^{h}$ denote the latitude and longitude of satellite $s_k$, respectively. The geocentric angle $\varphi$ between the two satellites is given by the spherical law of cosines:
\begin{equation}
  \varphi = \arccos\bigl[\sin\theta_{1}^{v}\sin\theta_{2}^{v} + \cos\theta_{1}^{v}\cos\theta_{2}^{v}\cos(\theta_{1}^{h} - \theta_{2}^{h})\bigr],
  \label{eq:geocentric}
\end{equation}
The two satellites are mutually LoS visible if and only if their geocentric angle does not exceed the sum of their individual horizon half-angles, i.e.,
$\varphi \leq\rho_1 + \rho_2$.

\begin{figure}[!t]
  \centering
  \includegraphics[width=0.68\columnwidth]{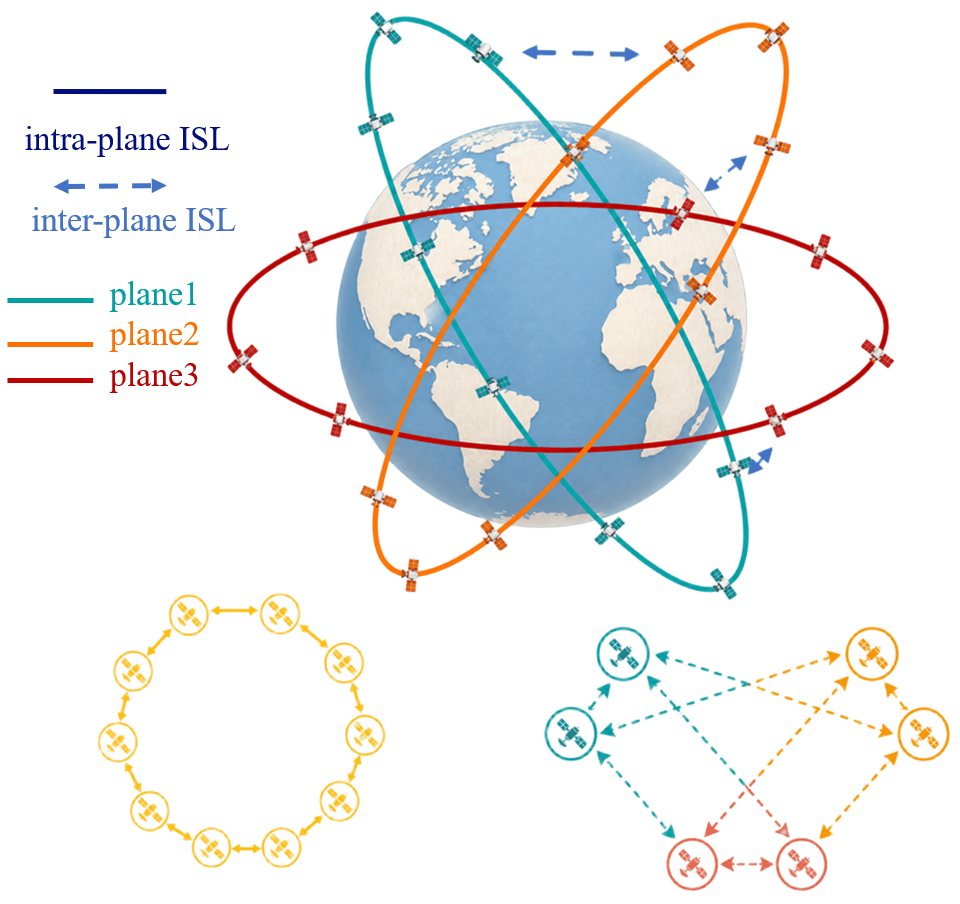}
  \caption{System model of the LEO satellite constellation.}
  \label{fig:constellation}
\end{figure}

\textit{2) Doppler reliability}:
The Doppler shift captures the dynamic dependency of ISL reliability. A clear LoS is insufficient to ensure link usability under high relative velocity. The Doppler shift between two satellites is $f_d = \psi f_c / c$, where $\psi$ is the relative speed, $f_c$ is the carrier frequency, and $c$ is the speed of light. The two satellites are considered Doppler reliable if and only if $f_d \leq f_{\max}$.

Once both metrics are satisfied, the satellite pair is considered credible, enabling successful information transmission between the two satellites.

\subsection{Satellite Federated Learning Model}

Assuming each satellite $s_{i,j}$ holds a local dataset $\mathcal{D}_{i,j}$, all satellites collaboratively learn a global model by solving the following empirical risk minimization (ERM) problem:
\begin{equation}
  \min_{\mathbf{w} \in \mathbb{R}^d} \quad  f(\mathbf{w}) = \frac{1}{NK} \sum_{i=1}^{N} \sum_{j=1}^{K} f_{i,j}(\mathbf{w}),
\end{equation}
where
\begin{equation}
  f_{i,j}(\mathbf{w}) = \frac{1}{|\mathcal{D}_{i,j}|} \sum_{\mathbf{z} \in \mathcal{D}_{i,j}} \ell(\mathbf{w}; \mathbf{z}),
\end{equation}
and $\mathbf{w}$ denotes the model parameters with dimension $d$, $\ell(\cdot; \mathbf{z})$ is the loss on sample $\mathbf{z}$. The operator $|\cdot|$ denotes the cardinality of the set.
Inspired by~\cite{ref:flsnn}, we adopt a constellation-aware federated learning architecture that combines intra-plane ring all-reduce with inter-plane gossip averaging, ensuring efficient and stable model aggregation under the physical constraints of the satellite network. The learning procedure at each global iteration $r$ proceeds in the following four steps, as shown in Fig~\ref{fig:constellation}:

\textbf{Step 1 --- Model Distribution.}
At the beginning of iteration $r$, each orbital plane $i$ holds a local model copy $\mathbf{w}_i^r$. Within each plane, the $K$ satellites are organized as a logical ring topology based on their stable relative positions along the orbit. In this ring, each satellite $s_{i,j}$ has a unique predecessor $s_{i,j-1}$ and a unique successor $s_{i,j+1}$ (indices modulo $K$). The satellite that participated in the previous inter-plane aggregation disseminates $\mathbf{w}_i^r$ to all other satellites $s_{i,j}$ in the same orbit by forwarding it along this ring, using the reliable intra-plane ISLs. After this step, every satellite in orbital plane $i$ holds $\mathbf{w}_i^r$ as its initial model for the current round, and the ring topology used for distribution also serves as the communication structure for subsequent aggregation.


\textbf{Step 2 --- Local Training.}
Each satellite $s_{i,j}$ updates its local model with a mini-batch optimizer AdamW~\cite{adamw}.The learning-rate schedule depends on the global round and is shared by all satellites, whereas the optimizer moments remain local.
We denote the weights after local training by $\mathbf{w}_{i,j}^{r+\frac{1}{2}}$.


\textbf{Step 3 --- Intra-Plane Model Aggregation (Ring All-Reduce).}
Using the same ring topology established in Step 1, the $K$ satellites perform ring all-reduce for intra-plane aggregation. Each satellite $s_{i,j}$ sends its local update $\mathbf{w}_{i,j}^{r+\frac{1}{2}}$ to its successor $s_{i,j+1}$ and receives an update from its predecessor $s_{i,j-1}$. After $K$ such communication rounds, every satellite in the plane obtains the exact average of all local models within that orbit:
\begin{equation}
  \label{eq:intra-plane}
  \mathbf{w}_i^{r+\frac{1}{2}} = \frac{1}{K} \sum_{j=1}^{K} \mathbf{w}_{i,j}^{r+\frac{1}{2}}, \quad \forall i \in \{1, \dots, N\}.
\end{equation}
At this point, each orbit $i$ has a unified intra-plane model $\mathbf{w}_i^{r+\frac{1}{2}}$.

\textbf{Step 4 --- Inter-Plane Model Aggregation (Gossip Averaging).}
Unlike the stable intra-plane ring topology, inter-plane ISLs are intermittent due to varying inclinations, altitudes, and relative velocities, making it impractical to achieve global aggregation across all $N$ orbits in a single round. Instead, a gossip averaging scheme is adopted: each orbital plane $i$ communicates only with its immediate neighbors on a pre-defined inter-plane connectivity graph $\mathcal{G} = (\mathcal{V}, \mathcal{E})$, where $\mathcal{V} = \{1, \dots, N\}$ and $e_{i,i'} \in \mathcal{E}$ only if there exists at least one credible pair between orbits $i$ and $i'$ (i.e., satisfying both LoS visibility and Doppler reliability constraints). In each gossip round, orbit $i$ exchanges its model $\mathbf{w}_i^{r+\frac{1}{2}}$ with its neighbors $\mathcal{N}_i = \{i' \mid e_{i,i'} \in \mathcal{E}\}$ and updates \cite{ref:flsnn}:
\begin{equation}
  \label{eq:gossip}
  \mathbf{w}_i^{r+1} = (1 - \alpha) \mathbf{w}_i^{r+\frac{1}{2}} + \frac{\alpha}{|\mathcal{N}_i|} \sum_{i' \in \mathcal{N}_i} \mathbf{w}_{i'}^{r+\frac{1}{2}},
\end{equation}
where $\alpha \in (0, 1)$ is the gossip mixing weight. Through repeated gossip rounds across iterations, model information gradually diffuses across the entire constellation, reaching consensus over time. This avoids the high communication overhead of global aggregation while remaining robust to the intermittent nature of inter-plane ISLs.

\section{Proposed Orbital Attention Leaky Integrate-and-fire (\OrbitALIF)  Framework}\label{sec:framework}

\subsection{Motivation and Architectural Overview}
\label{sec:framework-overview}
\begin{figure}[!t]
  \centering
  \includegraphics[width=0.8\columnwidth]{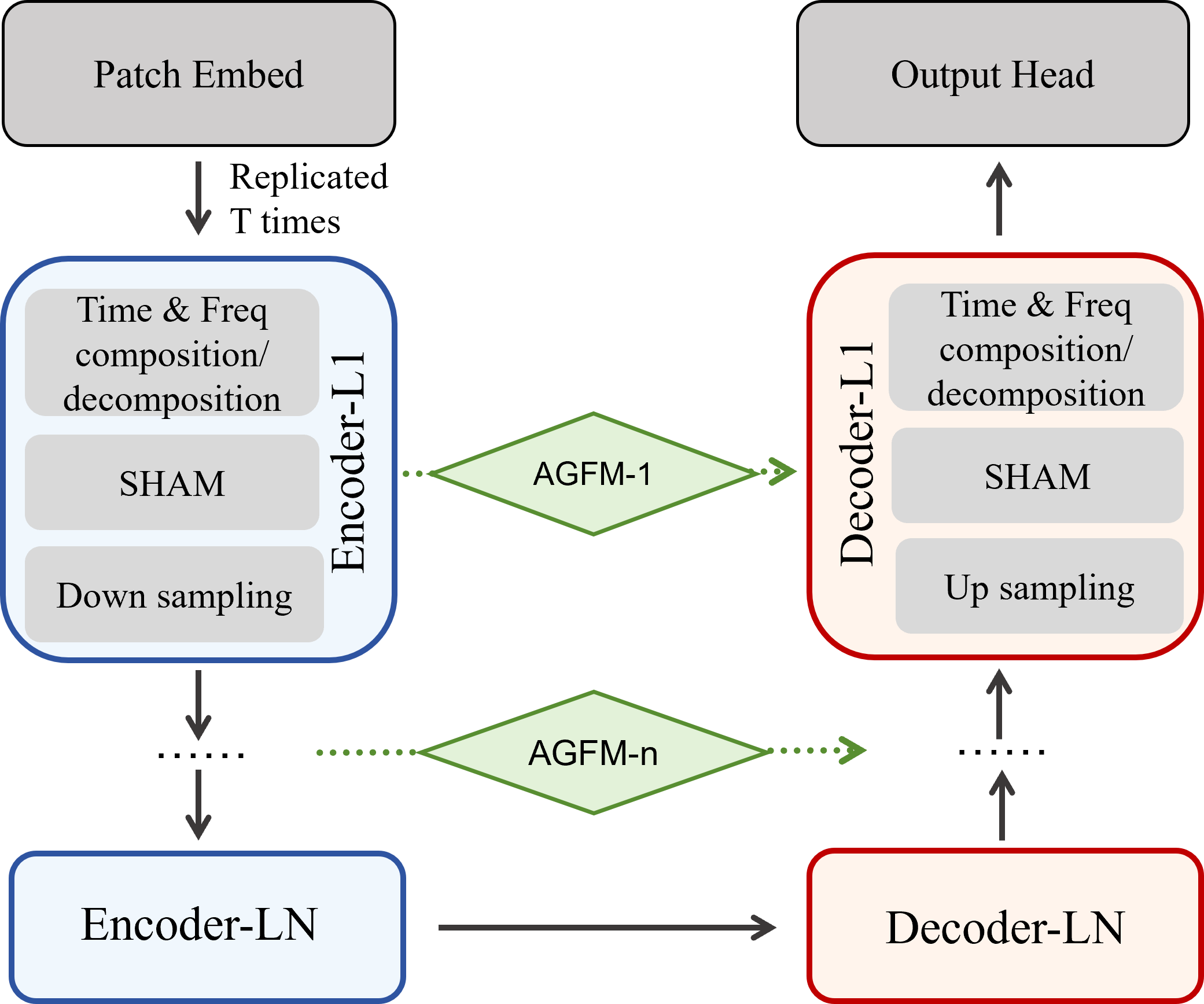}
  \caption{The core Encoder---Decoder architecture of the proposed \OrbitALIF\ framework.}
  \label{fig:SNNmodel}
\end{figure}

The system model in Section~\ref{sec:sysmodel} reveals three constraints for onboard cloud removal: tight power budgets,
intermittent ISLs, and heterogeneous data across orbital planes.
These call for a lightweight yet expressive backbone that extracts rich features under limited computation and generalizes across diverse conditions.

To meet these constraints, we propose the orbital attention leaky integrate-and-fire (\OrbitALIF) framework, as shown in Fig~\ref{fig:SNNmodel}, which features a U-shaped architecture based on the widely adopted U-Net~\cite{Unet}.
The input cloudy image
$\mathbf{c}\in\mathbb{R}^{3\times H\times W}$ is embedded and
replicated along a synthetic time axis with $T$ time steps, enabling
the network to process both spatial and temporal information
simultaneously. Building upon this design, we introduce two key
modules tailored for onboard cloud removal:

\begin{itemize}
  \item \textbf{Adaptive Gated Fusion Module (AGFM)}. In a standard
        U-Net, encoder features are directly concatenated or summed with
        decoder features at the same resolution level. However, when clouds
        are dense or observations from different satellites vary
        significantly, the encoder features may contain information that is
        not perfectly aligned with the decoder's current representation.
        AGFM replaces this rigid connection with a learnable gating
        mechanism, allowing the network to adaptively decide how much
        information to fuse at each spatial location.
  \item \textbf{Spectral-Spatial Hybrid Attention Module (SHAM)}.
        Inspired by~\cite{ref:fstasnn2025}, this module enhances the backbone
        by jointly attending to the temporal and frequency domains. Different time steps of the spiking rollout carry unequal information as the membrane state accumulates, and cloud artifacts occupy characteristic spatial-frequency ranges. SHAM learns to emphasize informative components in both domains.
\end{itemize}

The final prediction is formed by a global residual connection:
\begin{equation}
  \hat{\mathbf{y}} = \mathbf{c} +
  \mathrm{Conv}_{3\times3}\!\Bigl(
  \tfrac{1}{T}\textstyle\sum_{t=1}^{T}\mathbf{s}[t]\Bigr),
  \label{eq:globalresidual}
\end{equation}
where $\hat{\mathbf{y}}\in\mathbb{R}^{3\times H\times W}$ is the
restored cloud-free image, $\mathbf{s}[t]\in\mathbb{R}^{C\times H
    \times W}$ is the processed spike feature map at time step $t$ with $C$ channels,
and $\mathrm{Conv}_{3\times3}$ maps the temporally averaged features to
3 RGB channels. The global residual formulation lets the network learn
only the cloud correction, reducing the optimization burden and
stabilizing training. The complete model contains only $2.30$\,M
parameters, making it suitable for resource-constrained satellite
deployment.

The two proposed modules are detailed in the following subsections. For further details regarding the backbone architecture, readers are referred to~\cite{ref:esdnet}.

\subsection{Adaptive Gated Fusion Module (AGFM)}
\label{sec:framework-agfm}

In a standard U-Net, the skip connection at each level directly
concatenates the encoder feature map with the corresponding decoder
feature map. This design assumes that encoder features at a given
resolution are always beneficial to the decoder at the same resolution.
In practice, however, encoder and decoder features can have
substantial semantic discrepancies since the encoder captures
increasingly abstract representations, while the decoder reconstructs
fine-grained spatial details. Under the non-IID data distributions
inherent to satellite constellations, these discrepancies become more
pronounced, and indiscriminate fusion can introduce noise rather than
useful information.

AGFM addresses this problem by inserting a lightweight learnable
gate between each encoder--decoder skip connection. Specifically, let
$\mathbf{e} \in \mathbb{R}^{C \times H \times W}$ be the encoder
feature and $\mathbf{d} \in \mathbb{R}^{C \times H \times W}$ be the
decoder feature at the same resolution level. The gated fusion is
performed as:
\begin{equation}
  \mathbf{f} = (1 - \mathbf{g}) \odot \mathbf{e} + \mathbf{g} \odot \mathbf{d},
  \label{eq:agfm}
\end{equation}
where $\odot$ denotes element-wise multiplication, and the gate $\mathbf{g}$ is
produced by applying a $1\times1$ convolution to the concatenated
features, followed by a sigmoid activation.
Unlike a single scalar gate, $\mathbf{g}$ is both per-pixel and per-channel:
it can suppress the encoder branch at heavily clouded pixels, where
the contaminated encoder feature may hurt the output, while preserving
it at clear pixels.

AGFM offers three advantages for onboard deployment: minimal parameter
overhead, improved representational capacity with negligible inference cost, and better generalization under heterogeneous
orbital data distributions.

\subsection{Spectral-Spatial Hybrid Attention Module (SHAM)}
\label{sec:framework-sham}

While the U-Net backbone processes spatial features effectively, cloud
removal demands reasoning across two additional dimensions: time and
frequency. 
In the temporal domain, different time steps contribute unequally to the final prediction and some steps may carry more useful cloud information than others. In the frequency domain, cloud artifacts
exhibit characteristic spectral signatures. Inspired by~\cite{ref:fstasnn2025}, SHAM jointly models information from the temporal and frequency domains.


\textbf{Temporal attention branch.}
Given the input feature tensor
$\mathbf{x}\in\mathbb{R}^{T\times C\times H\times W}$, this branch
computes the temporal attention vector as
\begin{equation}
\mathbf{a}
=
\sigma\!\left(
\mathrm{m}_{C}\!\left[
\mathrm{FC}\!\left(
\lambda_{1}\mathrm{AvgPool}(\mathbf{x})
+
\lambda_{2}\mathrm{MaxPool}(\mathbf{x})
\right)
\right]
\right),
\label{eq:temporal-attention}
\end{equation}
where $\lambda_1$ and $\lambda_2$ are learnable
coefficients initialized to $0.5$. Two pooling
operations aggregate spatial dimensions while fully connected layer operates along time dimension.
The resulting features are averaged over channels by
$\mathrm{m}_{C}(\cdot)$ and passed through the sigmoid function
$\sigma(\cdot)$, producing $\mathbf{a}\in(0,1)^T$. Each element 
$a_t$ rescales the feature map at simulation time step $t$.

\textbf{Spectral attention branch.}
This branch operates in the frequency domain to emphasize informative
spectral bands while preserving spatial structure. It first computes
the temporally averaged feature
$\bar{\mathbf{x}} = \frac{1}{T}\sum_t\mathbf{x}[t]$, then applies a
2-D real FFT to obtain the amplitude spectrum
$|\mathcal{F}(\bar{\mathbf{x}})|$ and the phase
$\angle\mathcal{F}(\bar{\mathbf{x}})$. The amplitude is reweighted via
a lightweight $1\times1$ multilayer perceptron (MLP), while the phase is preserved unchanged. Phase encodes the spatial location of edges and structures,
modifying it would distort image geometry. After inverse FFT, the
result is passed through a $7\times7$ convolution to produce a spatial
attention map $\mathbf{A}_s \in \mathbb{R}^{H \times W}$.

The final output of SHAM is given by
\begin{equation}
  \mathrm{SHAM}(\mathbf{x}) \;=\;
  \mathbf{x} + \sigma(\lambda_s)\cdot (\mathbf{a} + \mathbf{A}_s + \mathbf{A}_s \odot \mathbf{a})
  \odot \mathbf{x},
  \label{eq:sham}
\end{equation}
where $\mathbf{a}\in\mathbb{R}^{T}$ is broadcast along the channel and spatial dimensions, scaling each time step $t$ by
$[\mathbf{a}_t]_t$ across all channels and pixels. Meanwhile,
$\mathbf{A}_s\in\mathbb{R}^{H\times W}$ is broadcast along the time
and channel dimensions, scaling each spatial position $(h,w)$ by
$\mathbf{A}_s[h,w]$ across all time steps and channels.
$\lambda_s\in\mathbb{R}$ is a learnable scalar initialized to
zero. $\sigma(\cdot)$ denotes the sigmoid activation function.The residual formulation ensures training stability.

SHAM is well-suited for onboard training and inference for two reasons. First,
the 2-D FFT operation is highly optimized on modern processors and
adds low latency. Second, both the temporal MLP and the spectral
$1\times1$ MLP introduce negligible parameter overhead, preserving the
$2.30$\,M-parameter budget of the full model. By adaptively
weighting time steps and frequency components, SHAM enables the
network to extract richer representations from the same computational
budget, making it particularly effective for the diverse cloud
patterns encountered across satellite orbits.

\subsection{Training Objective}
\label{sec:framework-loss}

The entire \OrbitALIF{} network is trained with a combined Charbonnier--SSIM
loss~\cite{ref:charbonnier_ssim}. Specifically, the Charbonnier component penalizes pixel-wise reconstruction errors while remaining smooth near zero, whereas the SSIM component encourages consistency in local luminance and spatial structure, helping mitigate the over-smoothed textures. Their combination therefore provides complementary reconstruction constraints for preserving both fidelity and structural detail. The training objective is defined as

\begin{equation}
\mathcal{L}(\hat{\mathbf{y}},\mathbf{y})
=
\frac{1}{|\Omega|}\sum_{p\in\Omega}
\sqrt{(\hat{y}_p-y_p)^2+\varepsilon^2}
+
\lambda\left(1-\mathrm{SSIM}
(\hat{\mathbf{y}},\mathbf{y})\right),
\label{eq:loss}
\end{equation}
where the first term is Charbonnier loss and the second is SSIM loss, $\mathrm{SSIM}(\cdot,\cdot)\in[-1,1]$. $\Omega$ is the set of all pixels in a figure, $\hat{y}_p$ and
$y_p$ are the predicted and ground-truth values at pixel $p$,
$\lambda\!=\!0.1$ balances the two terms.

\section{Experiments}\label{sec:experiments}


\subsection{Centralized Evaluation and Ablation}
\label{sec:exp-centralized}
We first conduct centralized training on the CUHK-CR~\cite{diffusion}
dataset. All models are trained for $400$
epochs using AdamW ($\beta_1=0.9$, $\beta_2=0.999$, $\epsilon=10^{-8}$,
weight decay $0$).Batch size is $4$ for training and $1$ for testing. The learning rate is linearly warmed up over the first three epochs and then annealed by a cosine schedule from $10^{-3}$ to $10^{-7}$. 
\begin{table}[!t]
  \vspace{0.2cm}
  \centering
  \caption{Centralized cloud removal on CUHK-CR1\cite{diffusion}. Ops. denotes MACs for ANN models
    and SOPs for \OrbitALIF{}. w/o means ``without''.}
  \setlength{\tabcolsep}{3.5pt}
  \renewcommand{\arraystretch}{0.95}
  \footnotesize
  \begin{tabular}{lccc}
    \toprule
    Method                  & Params (M)      & Ops. (G)     & PSNR (dB) \\
    \midrule
    CVAE~\cite{Cave}        & 15.56           & 37.13        & 24.252    \\
    MemoryNet~\cite{Memory} & 3.64            & 548.65       & 26.073    \\
    \midrule
    \OrbitALIF\ w/o SHAM    & 2.22            & 3.5          & 24.854    \\
    \OrbitALIF\ w/o AGFM    & 2.30            & 3.7          & 24.774    \\
    \textbf{\OrbitALIF{} (ours)}
                            & \textbf{2.30}   & \textbf{3.7}
                            & \textbf{25.374}                            \\
    \bottomrule
  \end{tabular}

  \label{tab:cr1}
\end{table}
Table~\ref{tab:cr1} compares \OrbitALIF\ against two state-of-the-art
cloud-removal baselines. The first, \textbf{CVAE}~\cite{Cave},
employs a Vision Transformer (ViT)-based
encoder that maps the cloudy input to a latent distribution. The second,
\textbf{MemoryNet}~\cite{Memory}, introduces a three-granularity memory
augment layer combined with
contrastive learning to preserve and align deep features across
multi-stage construction.
\OrbitALIF\ achieves $25.374$\,dB PSNR with only $2.30$\,M parameters
and $3.7$\,G SOPs---the lowest operation count among all methods.
CVAE requires $6.8\times$ more parameters and $10.0\times$ more
operations yet yields $1.122$\,dB lower PSNR. MemoryNet attains the
highest raw PSNR ($26.073$\,dB) but at over $148\times$ the
computational cost, making it impractical for satellite deployment.
Ablation results further confirm the contribution of both proposed
modules: removing SHAM drops PSNR to $24.854$\,dB (with $2.22$\,M
parameters and $3.5$\,G SOPs) and removing AGFM drops it to
$24.774$\,dB (with $2.30$\,M parameters and $3.7$\,G SOPs), both
below $25$\,dB. Since AGFM only introduces a lightweight $1\times1$ gating
convolution at each skip connection, its parameter and operation
overhead is negligible.

\subsection{Decentralized Federated Learning and Energy Evaluation}
\label{sec:exp-energy}
We further evaluate \OrbitALIF\ under the decentralized federated
learning setup described in Section~\ref{sec:sysmodel}, using a
$5{\times}10$ Walker-Star constellation ($N=5$ orbital planes,
$K=10$ per plane, $50$ clients total). Client datasets are partitioned
from CUHK-CR1 and CR2 via Dirichlet sampling ($\beta=0.1$) over source labels, with at least $5$ samples per client.  The FL
protocol runs for $R=200$ global rounds,each comprising $E_{\text{intra}}=2$ intra-plane ring synchronizations, with $E_{\text{local}}=2$ local AdamW per client. 

To isolate the benefits of each design component, we construct two
baseline architectures.

\begin{itemize}\setlength\itemsep{0pt}
  \item \textbf{\OrbitANN}: Retains the full multi-domain architecture
        of \OrbitALIF{} (AGFM, SHAM) but replaces
        all LIF neurons with ReLU. This baseline quantifies the energy--accuracy
        trade-off introduced by spiking computation: the ANN variant
        avoids spiking sparsity but executes all operations at full
        precision, incurring the dense computational cost of the
        multi-domain design.
  \item \textbf{\OrbitUnet}: A plain U-Net \cite{unet2015} that serves as a
        traditional ANN baseline for cloud removal. It uses standard
        convolutional encoder--decoder skip connections without any
        attention or frequency-domain modules, matching the backbone capacity to that of \OrbitALIF.
\end{itemize}


\begin{figure}[!t]
  \centering
  \subfloat[PSNR Comparison]{%
    \includegraphics[width=0.88\columnwidth]{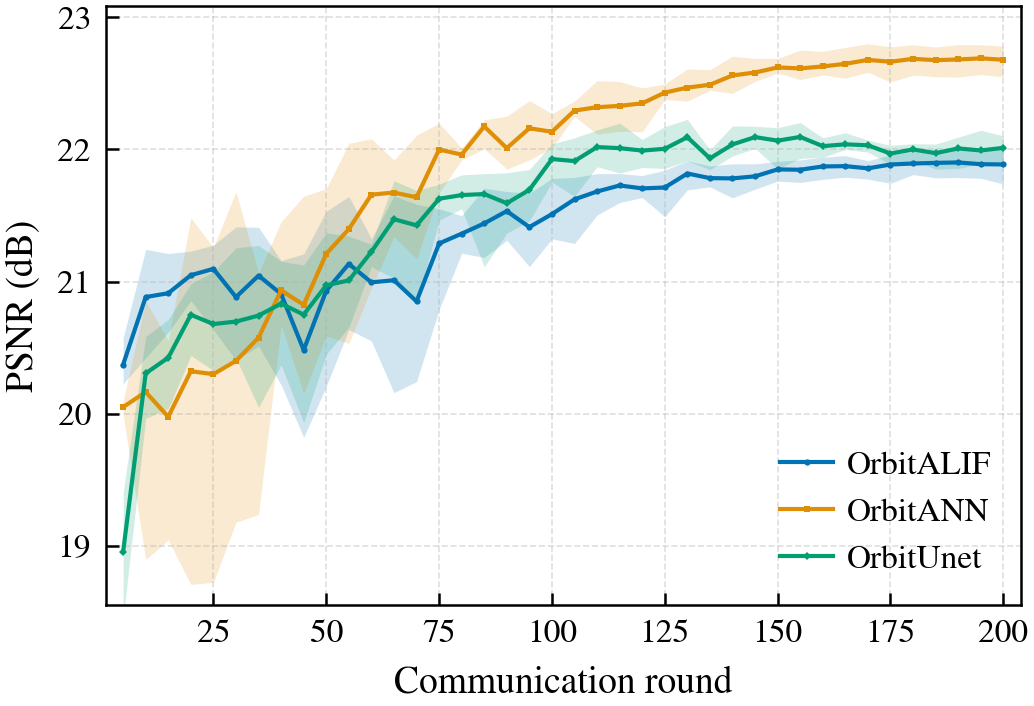}%
    \label{fig:psnr_comparsion}}
  \vspace{1mm}

  \subfloat[Energy Consumption]{%
    \includegraphics[width=0.88\columnwidth]{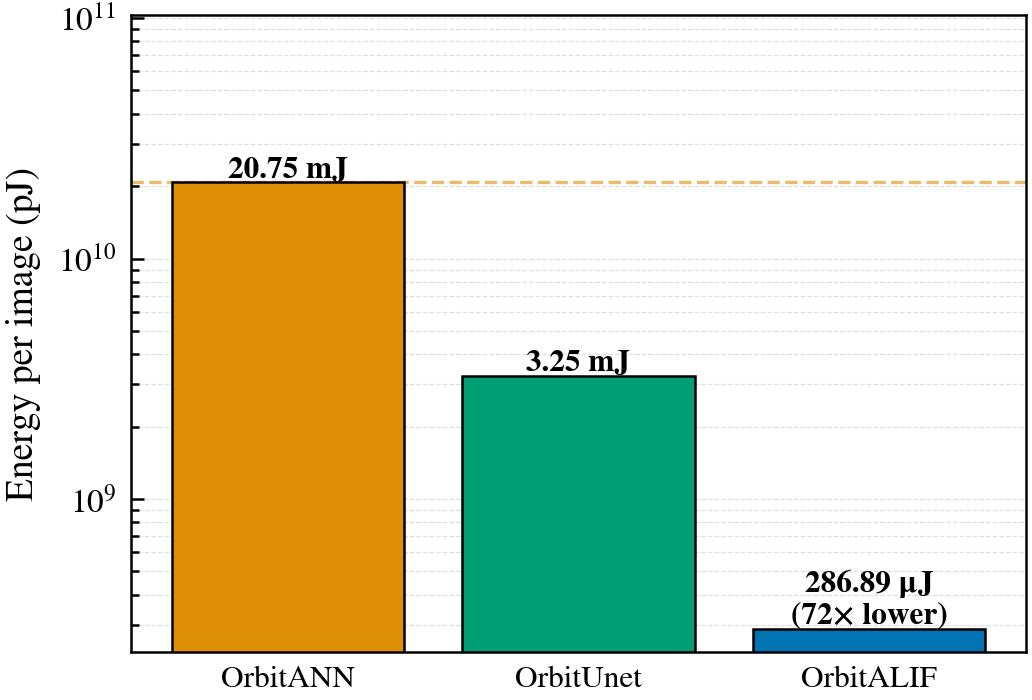}%
    \label{fig:energy_consumption}}
  \caption{Simulation results under the decentralized federated learning setup.
  (a) PSNR convergence curves. (b) Per-inference energy consumption.}
  \label{fig:simulation_results}
\end{figure}

Energy is estimated by attaching forward hooks to record the multiply-accumulate operations (MACs) for ANN models and the spike-driven synaptic operations (SOPs) for \OrbitALIF\ at each layer during inference, then multiplying the operation
count and input spike rate at each layer: $\text{SOP} =
  \text{MAC} \times \text{active rate}$. ANN consumption is computed
at $4.6$\,pJ/MAC (45\,nm CMOS~\cite{ref:horowitz2014}); SNN
consumption uses $77$\,fJ/SOP, consistent with spiking neuromorphic processor~\cite{77fj}
and prior works~\cite{77fjsnn}.

We first examine the PSNR convergence curves in
Fig.~\ref{fig:psnr_comparsion}. \OrbitANN\ ($22.679$\,dB) outperforms
\OrbitUnet\ ($22.012$\,dB) by $+0.667$\,dB, reflecting
the attribution of two key proposed modules. First, AGFM replaces the
rigid skip connections of a standard U-Net with a learnable
gating mechanism, enabling the network to adaptively balance
encoder and decoder features at each resolution level. Second,
SHAM jointly attends to both the temporal and frequency
domains, allowing the network to emphasize informative time
steps and frequency bands.
Meanwhile, \OrbitALIF\ ($21.886$\,dB) trails \OrbitANN\ by
$0.793$\,dB, reflecting the accuracy–efficiency trade-off introduced by spiking sparsity.

Turning to energy consumption in Fig.~\ref{fig:energy_consumption},
\OrbitALIF\ consumes only $0.287$\,mJ per inference on neuromorphic
hardware ($77$\,fJ/SOP), a $72.3\times$ ($98.6\%$) reduction compared
with \OrbitANN\ at $20.75$\,mJ ($4.6$\,pJ/MAC). Notably, \OrbitANN\
consumes more energy than \OrbitUnet, because the multi-domain
architecture (AGFM, SHAM) is designed and optimized for SNN rather
than ANN execution. When the same architecture is run as an ANN, all
operations execute unconditionally, leading to inflated energy
consumption. Despite this architectural mismatch, \OrbitALIF\ still
achieves an approximately $11\times$ energy reduction even when
compared with the \OrbitUnet\ baseline.
Figs.~\ref{fig:psnr_comparsion} and \ref{fig:energy_consumption}
demonstrate that the proposed spiking architecture delivers competitive
cloud removal quality with dramatically lower energy consumption,
validating that the brain-inspired design is well-suited for the tight
power budgets of LEO satellites.

\begin{figure}[!t]
  \centering
  \includegraphics[width=0.8\columnwidth]{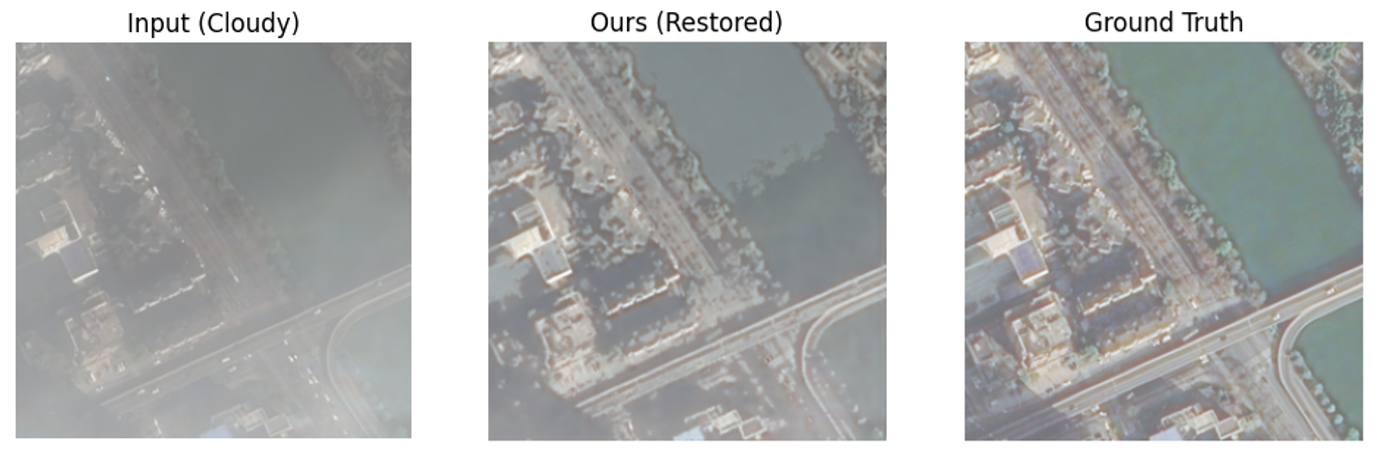}
  \caption{The cloud removal quality of \OrbitALIF\ under decentralized training.}
  \label{fig:restoration_performance}
\end{figure}

Fig.~\ref{fig:restoration_performance} further illustrates the visual
quality of \OrbitALIF\ under decentralized training. Despite operating
under the resource-constrained federated setting with only $2.30$\,M
parameters, the predicted images effectively recover cloud-free surface
details, confirming that the proposed framework achieves practical
cloud removal performance while remaining feasible for onboard
satellite deployment.

\section{Conclusion}\label{sec:conclusion}

We proposed \OrbitALIF, the first framework to perform both
onboard training and onboard inference for optical cloud removal
on a LEO satellite constellation. Rather than compressing an
existing ground-side model, we jointly designed the neuron activation,
construction network, and federated aggregation protocol from the
ground up for the constraints of Walker-Star operation.
Our evaluation reveals two key findings.
First, the brain-inspired spiking architecture achieves substantial energy savings, yielding a $72.3\times$ reduction in energy consumption relative to its ANN counterpart (\OrbitANN), at the cost of merely $0.793$ dB PSNR degradation. Second, the proposed SNN backbone outperforms the traditional U-Net (\OrbitUnet) by $+0.667$ dB in its ANN variant, confirming the effectiveness of the AGFM and SHAM modules for cloud removal. Furthermore, under centralized evaluation, the resulting spiking architecture \OrbitALIF\ achieves a competitive PSNR of $25.374$ dB while requiring only $2.30$ M parameters and $3.7$ G operations. These results collectively indicate that the lightweight design is well-suited for resource-constrained satellite deployment. Future work includes extending the framework to multi-spectral and
SAR-assisted inputs, and validating the energy bracket with hardware-measured power on Loihi-2, Speck, or Akida substrates.

\bibliographystyle{IEEEtran}
\bibliography{references}

\end{document}